\documentclass[11pt,a4paper]{article}
\usepackage[hyperref]{acl2021}
\usepackage{times}
\usepackage{latexsym}
\usepackage{booktabs}
\usepackage{graphicx}
\usepackage{url}
\usepackage{multirow}

\usepackage{microtype}

\aclfinalcopy 

\title{Nora: The Well-Being Coach}
\author{Genta Indra Winata$\thanks{\hspace{1mm} These two authors contributed equally.}$ $\hspace{0.45mm}^{1}$, Holy Lovenia$^{*1}$, Etsuko Ishii$^{1}$, Farhad Bin Siddique$^2$, \\\textbf{Yongsheng Yang}$^2$, \textbf{Pascale Fung}$^{1,2}$ \\ $^{1}$Center for Artificial Intelligence Research (CAiRE)\\
The Hong Kong University of Science and Technology\\
$^{2}$EMOS Technologies Inc.\\
\small\texttt{\{giwinata, eishii\}@connect.ust.hk, pascale@ece.ust.hk}}

\date{}

\begin{document}
\maketitle
\begin{abstract}
The current pandemic has forced people globally to remain in isolation and practice social distancing, which creates the need for a system to combat the resulting loneliness and negative emotions. In this paper we propose Nora, a virtual coaching platform designed to utilize natural language understanding in its dialogue system and suggest other recommendations based on user interactions. It is intended to provide assistance and companionship to people undergoing self-quarantine or work-from-home routines. Nora helps users gauge their well-being by detecting and recording the user's emotion, sentiment, and stress. Nora also recommends various workout, meditation, or yoga exercises to users in support of developing a healthy daily routine. In addition, we provide a social community inside Nora, where users can connect and share their experiences with others undergoing a similar isolation procedure. Nora can be accessed from anywhere via a web link and has support for both English and Mandarin.
\end{abstract}

\section{Introduction}
It is broadly accepted that loneliness poses a significant mental health problem~\cite{mushtaq2014relationship, beutel2017loneliness} and that surrounding social conditions of a person contributes to loneliness~\cite{tiwari2013loneliness}.
Lockdowns have been enforced worldwide to deal with the COVID-19 pandemic, causing many people to feel socially isolated. 
Many also have even been forced to self-quarantine for weeks after crossing a border or having close contact with a patient.
The resulting social isolation can have a negative impact on our mental health, and mental support for those under isolation is suggested~\citep{choi2020depression, zhao2020social}. 

\begin{figure}[!ht]
    \centering
    \includegraphics[scale=0.12]{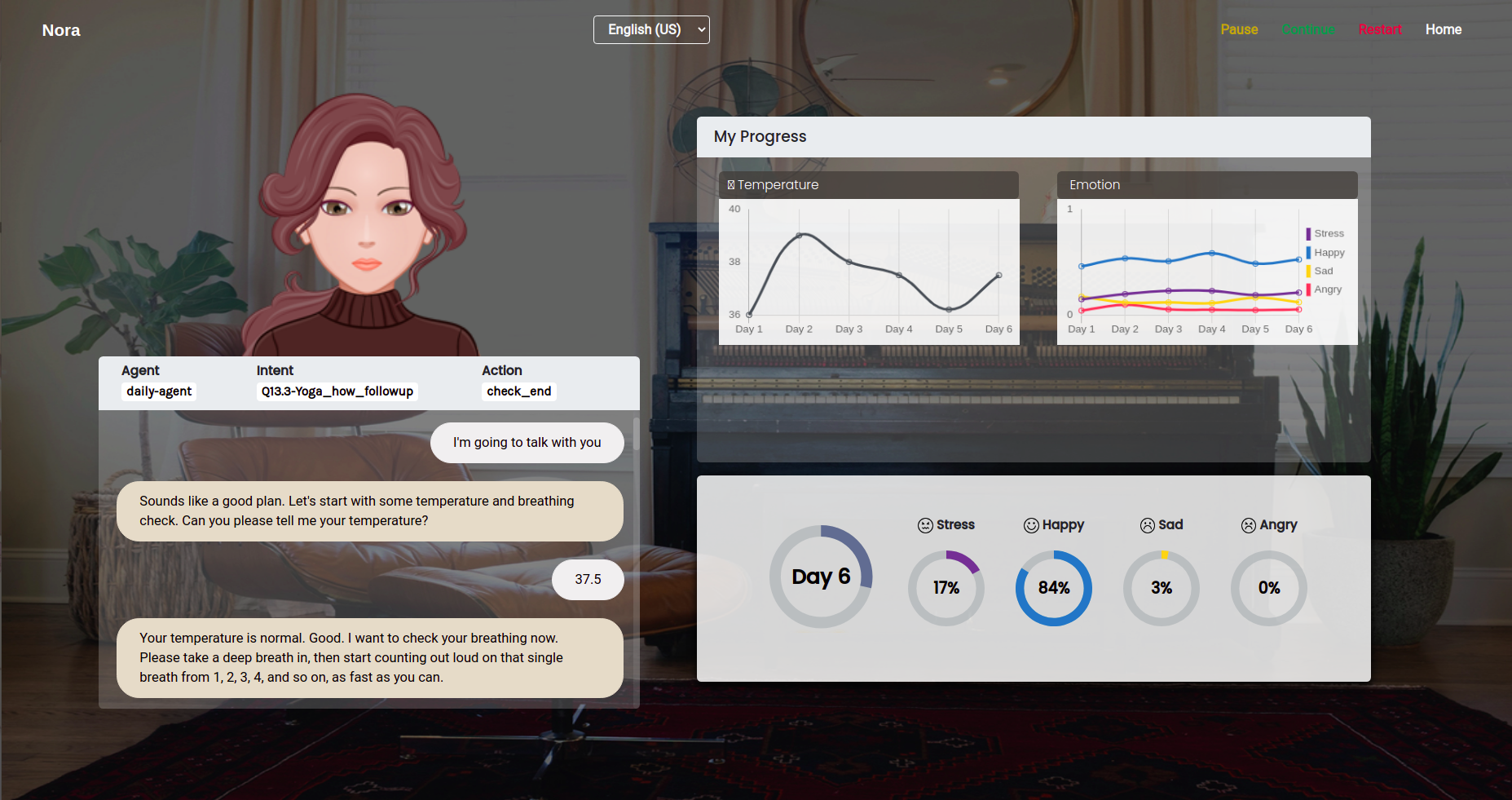}
    \caption{The user interface of the daily session in Nora.}
    \label{fig:daily-session}
\end{figure}
Traditionally, conversational agents have played the role of therapist, psychologist, or counselor to support mental health~\citep{weizenbaum1966eliza, cameron2017towards}.
While some are intended to replace professionals by diagnosing or providing a counselling~\citep{rudra2012escap, vaidyam2019chatbots}, others offer a more casual conversation to maintain mental well-being and encourage users to take formal sessions if necessary~\citep{rizzo2011simcoach, lee2020ihear, fitzpatrick2017delivering}.

\begin{figure*}[!ht]
    \centering
    \includegraphics[scale=0.12]{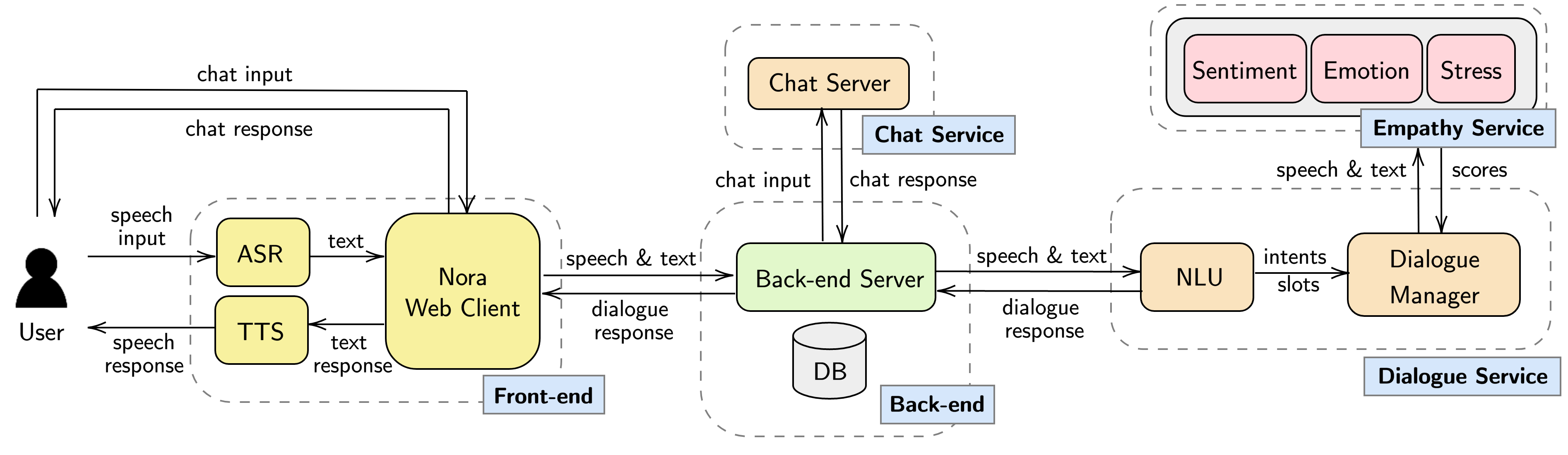}
    \caption{Nora Architecture. The system consists of front-end, dialogue service, empathy API, and chat server.}
    \label{fig:architecture}
\end{figure*}

Here, we introduce Nora~\footnote{Demo video: 
\url{https://youtu.be/M2FFer2GXe8} Website: \url{https://nora2.emos.ai} (supports Google Chrome Browser.)}, the well-being coach, extending the idea of an empathetic dialogue system that mimics a conversation with a psychologist~\citep{winata2017nora}.
To meet the emerging need caused by the COVID-19 pandemic, our system is currently optimized for people under self-quarantine.
Supporting English and Chinese, Nora asks a set of questions to screen for stress, depression, and health conditions, such as body temperature or shortness of breath during a conversation session.
The screened health conditions and emotional states are recorded throughout the self-quarantine.
We also implement user-to-user interaction functions so a user can have a chitchat with other users. This is inspired by online support groups for people with depression where they can seek information and support from others with a similar condition~\citep{griffiths2012effectiveness}. In this work, we demonstrate the possibility of combining affecting computing research in a practical scenario. It has the benefits of gauging and helping the users improve their psychological and physical well-being.

\section{System Description}
In this section, we describe the components of Nora and how Nora interacts with the user. Figure~\ref{fig:architecture} shows the overall architecture of the Nora system. Nora web client is the front-end that is shown to the user. The user can interact with the system by providing speech and text input. The speech input is used in the Nora session, while the text input is used in the chat feature.

The user speech is first converted to text by our automatic speech recognition (ASR). We use the ASR with a particular language based on the user preference. Then, the generated text is passed to the \textit{dialogue service}. The dialogue module identifies the user's intent and captures slot values by the natural language understanding (NLU) component. Then, the dialogue manager uses the intent and slots to plan to decide the response to the user according to the sentiment, emotion, and stress scores. Finally, to create a natural response, we output the speech by translating the text to speech using the text-to-speech (TTS) module. We use the TTS with the same language as the ASR. 

\subsection{Front-end Application}
We develop a web-based system by extending the interface design of the previous version of Nora~\cite{winata2017nora}. To access Nora, the user has to log in to the system by using social media accounts or registering a new account in the system. The user then will have an individual account, where they use to save the session logs and progress of all sessions. 
This will help the user to understand their mood progression throughout the program. In the front-end, we use English and Mandarin ASR and TTS to recognize speech to text and later converted the generated text to speech. 

\subsubsection{Nora Dashboard}
\begin{figure}[!ht]
    \centering
    \includegraphics[scale=0.12]{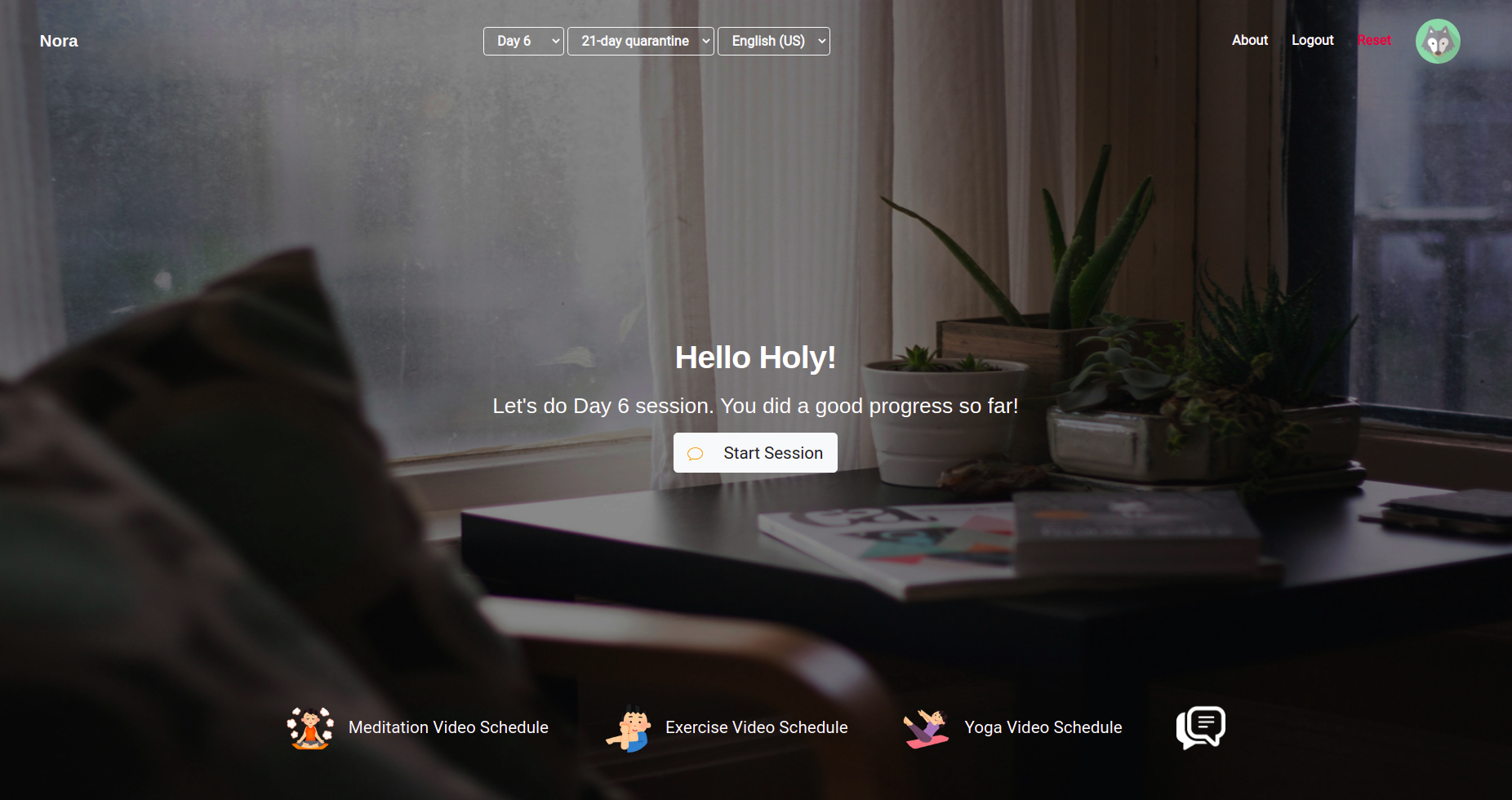}
    \caption{The user interface of the Nora dashboard page.}
    \label{fig:dashboard}
\end{figure}

We show the Nora dashboard page in Figure~\ref{fig:dashboard}. The dashboard page gives the user access to customize the user profile, preferences on meditation, exercise, yoga videos, and chat features. The user can also change their quarantine program and language preference on this page.

\subsubsection{Nora Console}
Figure~\ref{fig:daily-session} shows the daily session user interface design. On the left-hand side of the interface, the console displays the conversation box where we can follow the dialogue turns between the user and system. The system accepts speech input from the user to make the interaction as natural as possible. On the upper-right side, we show the progress bars, indicating the dynamics of temperature and empathetic scores across days. On the lower-right side, we display the session day indicator and empathetic scores, such as stress and emotion scores.

\subsubsection{Activity Recommendation}
\begin{figure}[!ht]
    \centering
    \includegraphics[scale=0.12]{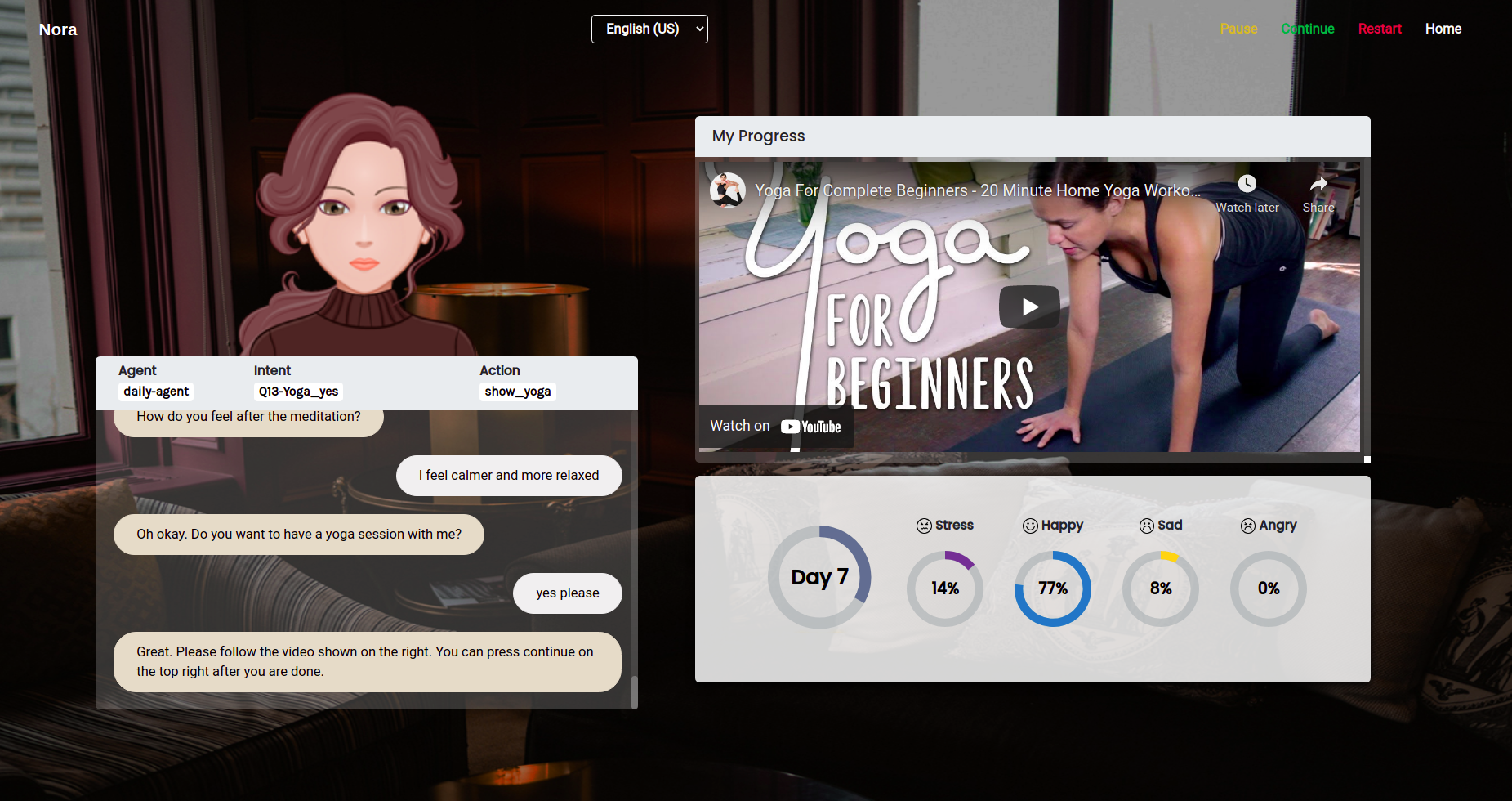}
    \caption{Example of a yoga session.}
    \label{fig:yoga-session}
\end{figure}

Nora also shows recommendations to the user in the console. For example, Nora can ask the user whether they want to do an exercise or a meditation, or a yoga session. If the user agrees to participate in the session, then Nora will show a window on the right-hand side of the interface as shown in Figure~\ref{fig:yoga-session}. After the session, the user can resume the conversation by clicking the ``continue" button.

\subsection{Back-end Server}
The back-end server manages the communication between the front-end and the other two services, which are chat service and dialogue service. To decouple the whole system, we set up the system into micro-services where they are independent, and each of them can be scaled individually. Specifically, in the back-end server, we store all user and dialogue information in the database. We use MongoDB to store non-structured dialogues, chat logs, and user setting data.

\subsection{Dialogue Service}
The dialogue service consists of NLU and dialogue manager modules. The NLU module predicts the user intent and slots related to the intent (e.g., user input: ``I am very grateful because of my parents", intent: ``grateful\_family", slot: ``parents"). Then, the intent and slots are used in the dialogue manager to generate appropriate responses back to the user by also conditioning on the sentiment and emotion scores from the empathy service.

\begin{figure*}[!t]
    \centering
    \includegraphics[scale=0.1]{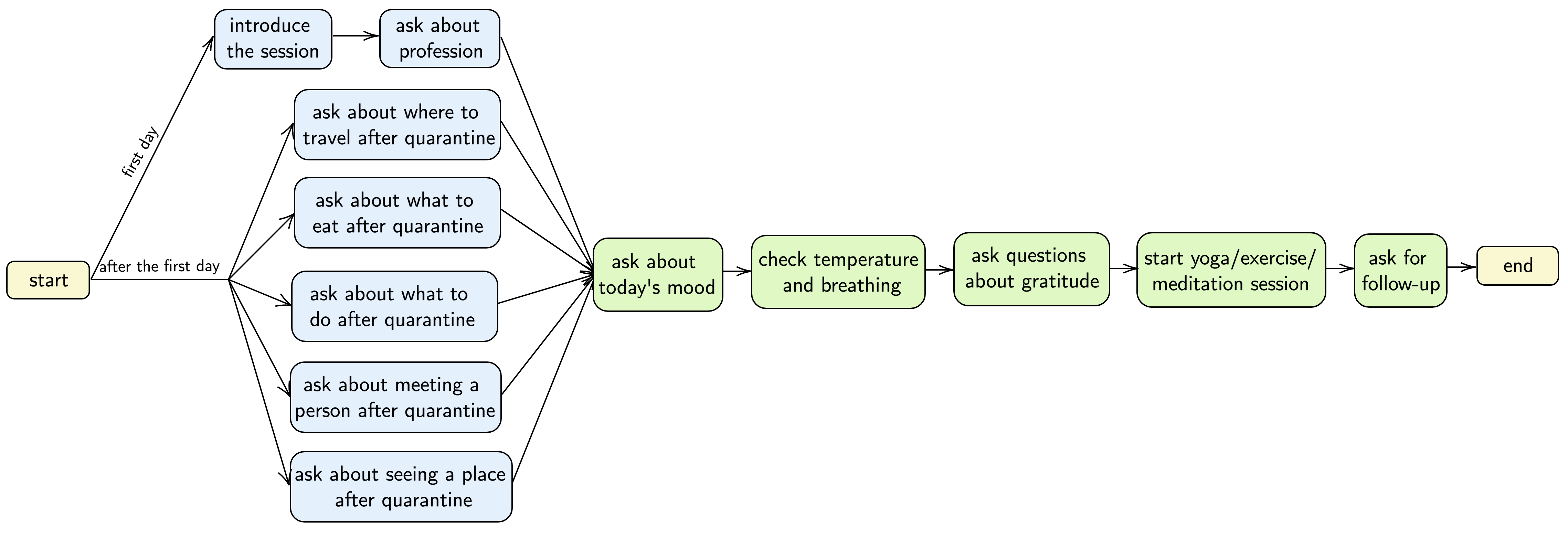}
    \caption{Dialogue Flow Design. We omit the information about the intent and slots for brevity.}
    \label{fig:dialogue-manager}
\end{figure*}

\subsection{Empathy Service}
The empathetic service consists of three APIs: sentiment, emotion, and stress. All of the APIs are run in our in-house system. For the sentiment analysis, we use a text based model that outputs a binary classification result (positive or negative) for each text. We train a Long short-term memory (LSTM) model using Word2Vec word embedding vectors \cite{mikolov2013distributed} as input for the text, and we train on the Movie Reviews dataset\footnote{\url{https://ai.stanford.edu/~amaas/data/sentiment/}} and we further improve our performance by training on the larger 1.6M labeled tweets from the Sentiment 140 dataset\footnote{\url{https://www.kaggle.com/kazanova/sentiment140}}. For the emotion classification models, we use text-based and audio-based emotion classification models. For our text-based model we use DeepMoji \cite{felbo2017using}, and for our audio based model we build a Convolutional Neural Network (CNN) model that takes raw audio as input, trained for multiple emotion classes. The model and dataset used for audio emotion is described in ~\citet{bertero2016real}, and we fuse the emotion scores from both text and audio using a weighted average method to get the final score for each emotion class. For the stress detection, we use a text-based stress detection model using BERT~\cite{devlin2019bert} to generate a stress score in a scale from 0 to 1 using the stress dataset from~\citet{winata2018attention}.\footnote{\url{https://github.com/gentaiscool/lstm-attention}} 

\section{Dialogue Design}
In this section, we describe the flow of our dialogue in the Nora session.
\subsection{Dialogue Flow}
Currently, we design the session in an agent-initiative interaction. Figure~\ref{fig:dialogue-manager} shows the dialogue flow of the Nora session from the start towards the end of the conversation. Several scenarios are depending on the day of the quarantine. Each day, the user will experience different interactions with Nora to avoid repetitions that may cause boredom.
\paragraph{First Day Conversation} 
During the first day, Nora introduces the session and what the user shall expect at the end of the session. Nora also asks the users to introduce themselves. Then, Nora starts the daily agent.
\paragraph{After First Day Conversation} 
After the first day, Nora begins to ask personal questions about plans after quarantine, such as asking about where to travel, what to eat, or what to do after quarantine. We design this agent to help the user retain hope and look forward to their future after quarantine.
\paragraph{Daily Conversation}
In the daily agent, Nora starts by asking the user's mood to understand what the user feels today. Then, Nora checks the user's health condition by asking about the temperature and whether the user has shortness of breathing. In the following turns, Nora allows the user to remember what they feel grateful for on that day. After that, Nora recommends the user to have a yoga, exercise, or meditation session. At last, Nora asks a follow-up question to know the feedback from the user.

\subsection{Temperature and Shortness of Breath Test}
As monitoring symptoms is one of the most essential things during self-quarantine, Nora requests the user to take their temperature and shortness of breath test daily.
If the user report between 32 to 38-degree Celsius, Nora regards as normal temperature, as high if between 38 to 43, and else invalid number to ask again.
To screen the shortness of breath, Nora asks the user to count up loud in a single breath, ``one, two, three, ...'', and a follow-up question whether they feel out of breath or not.
If the user has a fever or feels shortness of breath, Nora displays a hotline number and recommends consulting a doctor.
Temperature and breathing states are recorded and traceable throughout the quarantine.

\subsection{Activity Recommendation}
During the session, Nora recommends the user participate in the exercise, yoga, or meditation session. This allows the user to have an activity that can be done together with the dialogue agent. Nora shows a video of activity on the user interface as shown in Figure~\ref{fig:yoga-session}. We also allow the user to dynamically set their preferred activity videos for each quarantine day. This feature can be accessed on the dashboard.

\section{User-to-User Social Interaction}
It is reported to be promising for recovery to have social interaction with people in the same predicament~\citep{griffiths2012effectiveness}.
In addition to Nora-to-user interaction, we provide a few functions to enable users to communicate with each other, aiming to offer a platform where users can share their emotions, struggles, tips, etc.
There are three interaction interface types: 1-on-1 messaging, topic threads, and topic threads video call.

\subsection{1-on-1 Messaging}
1-on-1 messaging allows the user to have a conversation with another user.
To start a conversation, the user can add another user as a friend using their alias. Once added, they will appear in the contact list. From this point onward, they can exchange text messages and communicate.
We also provide a report feature to protect users from inappropriate messages.

\begin{figure}[t]
    \centering
    \includegraphics[scale=0.275]{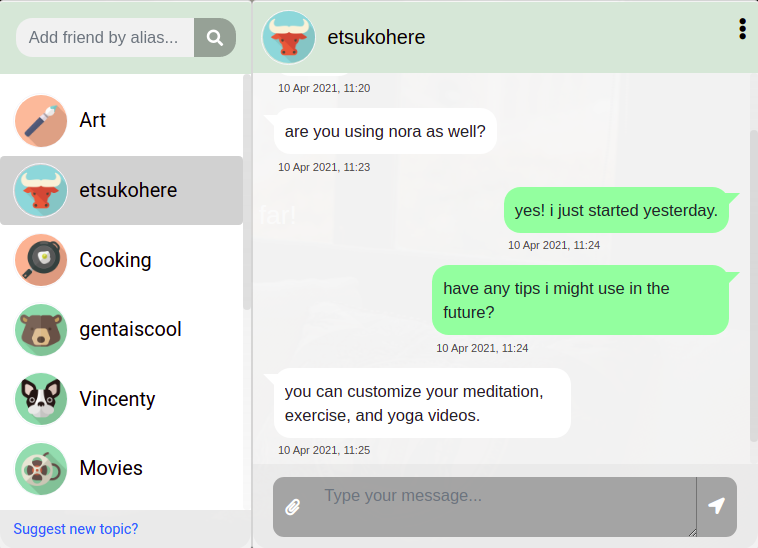}
    \caption{1-on-1 chat between users.}
    \label{fig:chat-101}
\end{figure}

There are a sender and a receiver for every message; both are registered as Firebase clients by the system. In our 1-on-1 chat implementation, while we store users' messages in our database, we utilize Firebase Cloud Messaging (FCM)\footnote{\url{https://firebase.google.com/docs/cloud-messaging/}} to send push notification from the server to the receiver's client-side every time there is a new message. After the notification is acquired on the receiver's client-side, it will generate a REST API call to the server to synchronize. Our server will fetch the new messages from the database and update the receiver's client-side according to the most recent chat log.

\subsection{Topic Threads}
We provide topics threads where users can freely and anonymously discuss predefined topics such as movies, cooking, music, etc.
Once a user identifies their interests, they will be automatically invited into relevant topic threads. 
In addition to text messages, topic threads also support a group video call feature, which will be explained in the following subsection.

After users update their profile with particular interests, their Firebase client will be subscribed to the related topics. Rather than the usual push notification, we use FCM topic messaging in topic threads. FCM topic messaging enables our server to send a push notification to multiple clients who have subscribed to a particular topic. When a user sends a message in the topic threads, other group members will receive a push notification and update their interface to load the new messages.

Users can decide not to partake in a topic thread anymore by removing the corresponding interest on their profile. When this happens, the server will fire an unsubscription request to Firebase so the user will not receive notifications from the topic thread again. Users can re-join the topic thread by adding the respective topic to their interest list later.

\begin{figure}[t]
    \centering
    \includegraphics[scale=0.275]{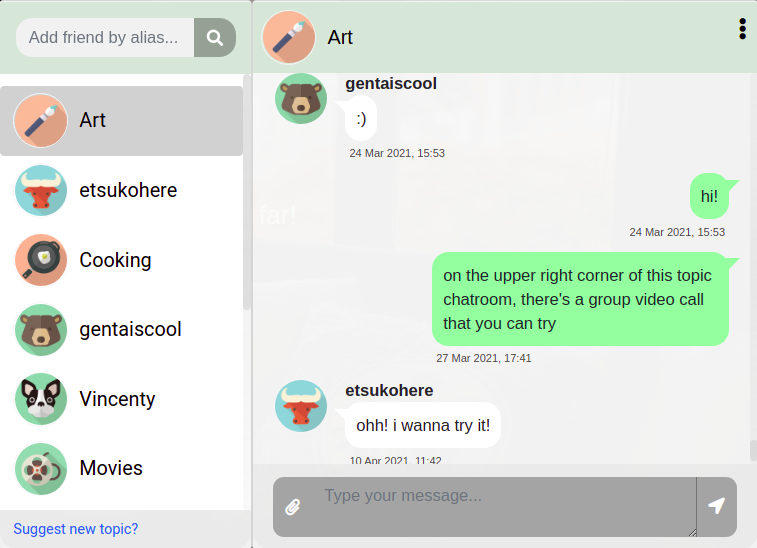}
    \caption{Interaction in a topic thread.}
    \label{fig:chat-topic}
\end{figure}

\subsection{Group Video Call}
To enrich the user-to-user interaction activity, we provide a group video call for each topic threads group.
There is an option on the upper right section of topic threads that can lead users to the relevant group video conference room when clicked. Members of the group can use it to have a face-to-face or a strict voice conversation without being pressured to reveal their original identities.

We implement the group video call feature using Zoom API's recurring meeting without a fixed time\footnote{\url{https://marketplace.zoom.us/docs/api-reference}}. The first-ever request to have a group video call in a topic thread makes an API call to Zoom to create a meeting and return the join URL to the user. The meeting credentials will then be saved in our database for future use. The subsequent requests will obtain those credentials from the database and retrieve the meeting via Zoom API. 

\section{Related Work}
\paragraph{Conversational Agents}
One of the major challenges in dialogue systems is to make systems respond in a more empathetic manner~\citep{zhou2018xiaoice, ma2020survey}, and several works have been explored to incorporate empathy in end-to-end open-ended chatbots~\citep{lin-etal-2019-moel, lin2020caire,madotto2020plug}. In the task-oriented dialogue system, ~\citet{lin2020mintl} explored the transfer learning approach to train end-to-end task-oriented dialogue system. ~\citet{lin2020xpersona} extended end-to-end conversational agents to the multilingual setting.

\paragraph{Companion Systems}
Recently, dialogue systems are attracting more interest in psychiatry, and a mental companion that offers guidance for self-help or mental well-being is often discussed as a promising application~\citep{vaidyam2019chatbots, bendig2019next}.
For example, SimCoach is aiming to empower military service members to seek professional resources if needed by providing pre-screening~\citep{rizzo2011simcoach}, Woebot offers a self-help program for students with anxiety and depression concern~\citep{fitzpatrick2017delivering}, and \citet{lee2020ihear} invented a chatbot which encourages self-disclosure.

\section{Conclusion}
In this paper we introduce Nora, a virtual coaching platform designed to help people undergoing mandatory quarantine or other self-isolation routines. Nora helps by gauging users' well-being with its in-built emotion, sentiment, and stress detection models, as well as by recommending various exercises (yoga, meditation, workout) based on their preferences. By interacting with Nora daily, users can set up a healthy daily routine and keep track of their mood and activities. Also, they can connect to other users going through a similar isolation procedure via the social community inside of Nora, where they can share their experiences and recommend exercises to each other. Nora also has support for both English and Mandarin languages, and can be accessed from anywhere via a browser and web-link. Such a system uses Human-Computer Interaction techniques together with the empathy services to help users in isolation improve their psychological and physical well-being, which is crucial in today's pandemic ridden world.

\section*{Acknowledgments}
We would thank Hubert Etienne, Zhaojiang Lin, Dan Su, Samuel Cahyawijaya, Andrea Madotto, and Yejin Bang for discussing the project. This work has been partially funded by ITF/319/16FP and MRP/055/18 of the Innovation Technology Commission, the Hong Kong SAR Government, and School of Engineering Ph.D.~Fellowship Award, the Hong Kong University of Science and Technology, and RDC 1718050-0 of EMOS.AI.

\bibliographystyle{acl_natbib}
\bibliography{anthology,acl2021}

\begin{thebibliography}{27}
\expandafter\ifx\csname natexlab\endcsname\relax\def\natexlab#1{#1}\fi

\bibitem[{Bendig et~al.(2019)Bendig, Erb, Schulze-Thuesing, and
  Baumeister}]{bendig2019next}
Eileen Bendig, Benjamin Erb, Lea Schulze-Thuesing, and Harald Baumeister. 2019.
\newblock \href {https://doi.org/10.1159/000501812} {The next generation:
  Chatbots in clinical psychology and psychotherapy to foster mental health –
  a scoping review}.
\newblock \emph{Verhaltenstherapie}, page 1–13.

\bibitem[{Bertero et~al.(2016)Bertero, Siddique, Wu, Wan, Chan, and
  Fung}]{bertero2016real}
Dario Bertero, Farhad~Bin Siddique, Chien-Sheng Wu, Yan Wan, Ricky Ho~Yin Chan,
  and Pascale Fung. 2016.
\newblock Real-time speech emotion and sentiment recognition for interactive
  dialogue systems.
\newblock In \emph{Proceedings of the 2016 conference on empirical methods in
  natural language processing}, pages 1042--1047.

\bibitem[{Beutel et~al.(2017)Beutel, Klein, Brähler, Reiner, Jünger, Michal,
  Wiltink, Wild, Münzel, Lackner, and et~al.}]{beutel2017loneliness}
Manfred~E. Beutel, Eva~M. Klein, Elmar Brähler, Iris Reiner, Claus Jünger,
  Matthias Michal, Jörg Wiltink, Philipp~S. Wild, Thomas Münzel, Karl~J.
  Lackner, and et~al. 2017.
\newblock \href {https://doi.org/10.1186/s12888-017-1262-x} {Loneliness in the
  general population: prevalence, determinants and relations to mental health}.
\newblock \emph{BMC Psychiatry}, 17(1).

\bibitem[{Cameron et~al.(2017)Cameron, Cameron, Megaw, Bond, Mulvenna, O'Neill,
  Armour, and McTear}]{cameron2017towards}
Gillian Cameron, David Cameron, Gavin Megaw, Raymond Bond, Maurice Mulvenna,
  Siobhan O'Neill, Cherie Armour, and Michael McTear. 2017.
\newblock \href {https://doi.org/10.14236/ewic/HCI2017.24} {Towards a chatbot
  for digital counselling}.
\newblock In \emph{Proc. BCS-HCI}, HCI '17, Swindon, GBR. BCS Learning and
  Development Ltd.

\bibitem[{Choi et~al.(2020)Choi, Hui, and Wan}]{choi2020depression}
Edmond Pui~Hang Choi, Bryant Pui~Hung Hui, and Eric Yuk~Fai Wan. 2020.
\newblock \href {https://doi.org/10.3390/ijerph17103740} {Depression and
  anxiety in hong kong during covid-19}.
\newblock \emph{International Journal of Environmental Research and Public
  Health}, 17(10):3740.

\bibitem[{Devlin et~al.(2019)Devlin, Chang, Lee, and
  Toutanova}]{devlin2019bert}
Jacob Devlin, Ming-Wei Chang, Kenton Lee, and Kristina Toutanova. 2019.
\newblock Bert: Pre-training of deep bidirectional transformers for language
  understanding.
\newblock In \emph{Proceedings of the 2019 Conference of the North American
  Chapter of the Association for Computational Linguistics: Human Language
  Technologies, Volume 1 (Long and Short Papers)}, pages 4171--4186.

\bibitem[{Felbo et~al.(2017)Felbo, Mislove, S{\o}gaard, Rahwan, and
  Lehmann}]{felbo2017using}
Bjarke Felbo, Alan Mislove, Anders S{\o}gaard, Iyad Rahwan, and Sune Lehmann.
  2017.
\newblock Using millions of emoji occurrences to learn any-domain
  representations for detecting sentiment, emotion and sarcasm.
\newblock In \emph{EMNLP}.

\bibitem[{Fitzpatrick et~al.(2017)Fitzpatrick, Darcy, and
  Vierhile}]{fitzpatrick2017delivering}
Kathleen~Kara Fitzpatrick, Alison Darcy, and Molly Vierhile. 2017.
\newblock \href {https://doi.org/10.2196/mental.7785} {Delivering cognitive
  behavior therapy to young adults with symptoms of depression and anxiety
  using a fully automated conversational agent (woebot): A randomized
  controlled trial}.
\newblock \emph{JMIR Ment Health}, 4(2):e19.

\bibitem[{Griffiths et~al.(2012)Griffiths, Mackinnon, Crisp, Christensen,
  Bennett, and Farrer}]{griffiths2012effectiveness}
Kathleen~M. Griffiths, Andrew~J. Mackinnon, Dimity~A. Crisp, Helen Christensen,
  Kylie Bennett, and Louise Farrer. 2012.
\newblock \href {https://doi.org/10.1371/journal.pone.0053244} {The
  effectiveness of an online support group for members of the community with
  depression: A randomised controlled trial}.
\newblock \emph{PLoS ONE}, 7(12).

\bibitem[{Lee et~al.(2020)Lee, Yamashita, Huang, and Fu}]{lee2020ihear}
Yi-Chieh Lee, Naomi Yamashita, Yun Huang, and Wai Fu. 2020.
\newblock \href {https://doi.org/10.1145/3313831.3376175} {"i hear you, i feel
  you": Encouraging deep self-disclosure through a chatbot}.
\newblock In \emph{Proc. CHI}, page 1–12, New York, NY, USA. ACM.

\bibitem[{Lin et~al.(2020{\natexlab{a}})Lin, Liu, Winata, Cahyawijaya, Madotto,
  Bang, Ishii, and Fung}]{lin2020xpersona}
Zhaojiang Lin, Zihan Liu, Genta~Indra Winata, Samuel Cahyawijaya, Andrea
  Madotto, Yejin Bang, Etsuko Ishii, and Pascale Fung. 2020{\natexlab{a}}.
\newblock Xpersona: Evaluating multilingual personalized chatbot.
\newblock \emph{arXiv preprint arXiv:2003.07568}.

\bibitem[{Lin et~al.(2019)Lin, Madotto, Shin, Xu, and
  Fung}]{lin-etal-2019-moel}
Zhaojiang Lin, Andrea Madotto, Jamin Shin, Peng Xu, and Pascale Fung. 2019.
\newblock \href {https://doi.org/10.18653/v1/D19-1012} {{M}o{EL}: Mixture of
  empathetic listeners}.
\newblock In \emph{Proceedings of the 2019 Conference on Empirical Methods in
  Natural Language Processing and the 9th International Joint Conference on
  Natural Language Processing (EMNLP-IJCNLP)}, pages 121--132, Hong Kong,
  China. Association for Computational Linguistics.

\bibitem[{Lin et~al.(2020{\natexlab{b}})Lin, Madotto, Winata, and
  Fung}]{lin2020mintl}
Zhaojiang Lin, Andrea Madotto, Genta~Indra Winata, and Pascale Fung.
  2020{\natexlab{b}}.
\newblock Mintl: Minimalist transfer learning for task-oriented dialogue
  systems.
\newblock In \emph{Proceedings of the 2020 Conference on Empirical Methods in
  Natural Language Processing (EMNLP)}, pages 3391--3405.

\bibitem[{Lin et~al.(2020{\natexlab{c}})Lin, Xu, Winata, Siddique, Liu, Shin,
  and Fung}]{lin2020caire}
Zhaojiang Lin, Peng Xu, Genta~Indra Winata, Farhad~Bin Siddique, Zihan Liu,
  Jamin Shin, and Pascale Fung. 2020{\natexlab{c}}.
\newblock Caire: An end-to-end empathetic chatbot.
\newblock In \emph{Proc. AAAI}, volume~34, pages 13622--13623.

\bibitem[{Ma et~al.(2020)Ma, Nguyen, Xing, and Cambria}]{ma2020survey}
Yukun Ma, Khanh~Linh Nguyen, Frank~Z. Xing, and Erik Cambria. 2020.
\newblock \href {https://doi.org/https://doi.org/10.1016/j.inffus.2020.06.011}
  {A survey on empathetic dialogue systems}.
\newblock \emph{Information Fusion}, 64:50 -- 70.

\bibitem[{Madotto et~al.(2020)Madotto, Ishii, Lin, Dathathri, and
  Fung}]{madotto2020plug}
Andrea Madotto, Etsuko Ishii, Zhaojiang Lin, Sumanth Dathathri, and Pascale
  Fung. 2020.
\newblock Plug-and-play conversational models.
\newblock In \emph{Proceedings of the 2020 Conference on Empirical Methods in
  Natural Language Processing: Findings}, pages 2422--2433.

\bibitem[{Mikolov et~al.(2013)Mikolov, Sutskever, Chen, Corrado, and
  Dean}]{mikolov2013distributed}
Tomas Mikolov, Ilya Sutskever, Kai Chen, Greg Corrado, and Jeffrey Dean. 2013.
\newblock Distributed representations of words and phrases and their
  compositionality.
\newblock \emph{arXiv preprint arXiv:1310.4546}.

\bibitem[{Mushtaq et~al.(2014)Mushtaq, Shoib, Shah, and
  Mushtaq}]{mushtaq2014relationship}
Raheel Mushtaq, Sheikh Shoib, Tabindah Shah, and Sahil Mushtaq. 2014.
\newblock \href {https://doi.org/10.7860/jcdr/2014/10077.4828} {Relationship
  between loneliness, psychiatric disorders and physical health ? a review on
  the psychological aspects of loneliness}.
\newblock \emph{Journal Of Clinical And Diagnostic Research}, 8(9):WE01--WE04.

\bibitem[{Rizzo et~al.(2011)Rizzo, Lange, Buckwalter, Forbell, Kim, Sagae,
  Williams, Difede, Rothbaum, Reger, and et~al.}]{rizzo2011simcoach}
Albert Rizzo, Belinda Lange, John~G. Buckwalter, Eric Forbell, Julia Kim, Kenji
  Sagae, Josh Williams, Joann Difede, Barbara~O. Rothbaum, Greg Reger, and
  et~al. 2011.
\newblock \href {https://doi.org/10.1515/ijdhd.2011.046} {Simcoach: an
  intelligent virtual human system for providing healthcare information and
  support}.
\newblock \emph{International Journal on Disability and Human Development},
  10(4).

\bibitem[{{Rudra} et~al.(2012){Rudra}, {Li}, and {Kavakli}}]{rudra2012escap}
Tarashankar {Rudra}, Manning {Li}, and Manolya {Kavakli}. 2012.
\newblock \href {https://doi.org/10.1109/HICSS.2012.249} {Escap: Towards the
  design of an ai architecture for a virtual counselor to tackle students' exam
  stress}.
\newblock In \emph{2012 45th Hawaii International Conference on System
  Sciences}, pages 2981--2990.

\bibitem[{Tiwari(2013)}]{tiwari2013loneliness}
Sarvadachandra Tiwari. 2013.
\newblock \href {https://doi.org/10.4103/0019-5545.120536} {Loneliness: A
  disease?}
\newblock \emph{Indian Journal of Psychiatry}, 55(4):320.

\bibitem[{Vaidyam et~al.(2019)Vaidyam, Wisniewski, Halamka, Kashavan, and
  Torous}]{vaidyam2019chatbots}
Aditya~Nrusimha Vaidyam, Hannah Wisniewski, John~David Halamka, Matcheri~S.
  Kashavan, and John~Blake Torous. 2019.
\newblock \href {https://doi.org/10.1177/0706743719828977} {Chatbots and
  conversational agents in mental health: A review of the psychiatric
  landscape}.
\newblock \emph{The Canadian Journal of Psychiatry}, 64(7):456--464.

\bibitem[{Weizenbaum(1966)}]{weizenbaum1966eliza}
Joseph Weizenbaum. 1966.
\newblock \href {https://doi.org/10.1145/365153.365168} {Eliza - a computer
  program for the study of natural language communication between man and
  machine}.
\newblock \emph{Commun. ACM}, 9(1):36–45.

\bibitem[{Winata et~al.(2017)Winata, Kampman, Yang, Dey, and
  Fung}]{winata2017nora}
Genta~Indra Winata, Onno Kampman, Yang Yang, Anik Dey, and Pascale Fung. 2017.
\newblock \href
  {https://www.isca-speech.org/archive/Interspeech_2017/pdfs/2050.PDF} {Nora
  the empathetic psychologist}.
\newblock In \emph{Proc. INTERSPEECH}, pages 3437--3438.

\bibitem[{Winata et~al.(2018)Winata, Kampman, and Fung}]{winata2018attention}
Genta~Indra Winata, Onno~Pepijn Kampman, and Pascale Fung. 2018.
\newblock Attention-based lstm for psychological stress detection from spoken
  language using distant supervision.
\newblock In \emph{Proc. ICASSP}, pages 6204--6208. IEEE.

\bibitem[{Zhao et~al.(2020)Zhao, Wong, Wu, Choi, Wang, and
  Lam}]{zhao2020social}
Sheng~Zhi Zhao, Janet Yuen~Ha Wong, Yongda Wu, Edmond Pui~Hang Choi, Man~Ping
  Wang, and Tai~Hing Lam. 2020.
\newblock \href {https://doi.org/10.3390/ijerph17186692} {Social distancing
  compliance under covid-19 pandemic and mental health impacts: A
  population-based study}.
\newblock \emph{International Journal of Environmental Research and Public
  Health}, 17(18):6692.

\bibitem[{Zhou et~al.(2018)Zhou, Gao, Li, and Shum}]{zhou2018xiaoice}
Li~Zhou, Jianfeng Gao, Di~Li, and Heung{-}Yeung Shum. 2018.
\newblock \href {http://arxiv.org/abs/1812.08989} {The design and
  implementation of xiaoice, an empathetic social chatbot}.
\newblock \emph{CoRR}, abs/1812.08989.

\end{thebibliography}


\end{document}